\def\year{2022}\relax
\documentclass[letterpaper]{article} 
\usepackage{aaai22}  
\usepackage{times}  
\usepackage{helvet}  
\usepackage{courier}  
\usepackage[hyphens]{url}  
\usepackage{graphicx} 
\usepackage{natbib}  
\usepackage{caption} 
\DeclareCaptionStyle{ruled}{labelfont=normalfont,labelsep=colon,strut=off} 
\usepackage{algorithm}
\usepackage{algorithmic}

\usepackage{newfloat}
\usepackage{listings}

\floatstyle{ruled}
\newfloat{listing}{tb}{lst}{}
\floatname{listing}{Listing}

\usepackage{booktabs}
\usepackage{amsmath}
\usepackage{xcolor}
\usepackage[colorlinks = True, citecolor = violet, linkcolor=blue, urlcolor = blue]{hyperref}
\usepackage{navigator}

\title{A Search Engine for Discovery of Scientific Challenges and Directions}
\author{
    Dan Lahav\textsuperscript{\rm 1} ~ Jon Saad Falcon\textsuperscript{\rm 2,4}~ Bailey Kuehl\textsuperscript{\rm 2}~ Sophie Johnson\textsuperscript{\rm 2}~ 
    Sravanthi Parasa\textsuperscript{\rm 6} \\ Noam  Shomron\textsuperscript{\rm 1}~ Duen Horng Chau\textsuperscript{\rm 4}~ Diyi Yang\textsuperscript{\rm 4} ~
    Eric Horvitz\textsuperscript{\rm 5} ~ Daniel S. Weld\textsuperscript{\rm 2,3}~ Tom Hope\textsuperscript{\rm 2,3}
}
\affiliations{
    \textsuperscript{\rm 1}Tel Aviv University ~ \textsuperscript{\rm 2}Allen Institute for AI ~ 
    \textsuperscript{\rm 3}University of Washington \\
    \textsuperscript{\rm 4}Georgia Institute of Technology ~
    \textsuperscript{\rm 5}Microsoft ~
    \textsuperscript{\rm 6}Swedish Medical Group
    
}

\begin{document}

\maketitle

\begin{abstract}

Keeping track of scientific challenges, advances and emerging directions is a fundamental part of research. However, researchers face a flood of papers that hinders discovery of important knowledge. In biomedicine, this directly impacts human lives. To address this problem, we present a novel task of extraction and search of scientific challenges and directions, to facilitate rapid knowledge discovery. We construct and release an expert-annotated corpus of texts sampled from full-length papers, labeled with novel semantic categories that generalize across many types of challenges and directions. We focus on a large corpus of interdisciplinary work relating to the COVID-19 pandemic, ranging from biomedicine to areas such as AI and economics. We apply a model trained on our data to identify challenges and directions across the corpus and build a dedicated search engine. In experiments with 19 researchers and clinicians using our system, we outperform a popular scientific search engine in assisting knowledge discovery. Finally, we show that models trained on our resource generalize to the wider biomedical domain and to AI papers, highlighting its broad utility. We make our data, model and search engine publicly available.\footnote{ \url{https://challenges.apps.allenai.org/}.}
\end{abstract}

\section{Introduction}

 \begin{quotation}
\noindent ``We can only see a short distance ahead, but we can see plenty there that needs to be done.''\\
\vspace{-0.05cm}
~~~~~~~~\quad\quad\quad\quad\quad\quad\quad\quad\quad\quad\quad \textit{-- Alan Turing, 1950}
 \end{quotation}

Success in scientific efforts hinges on identifying promising and important problems to work on, developing novel and effective solutions, and formulating hypotheses and directions for further exploration. Each new scientific advance helps address gaps in knowledge, including potential extensions and refinements of prior results. New advances often lead to new challenges and directions. With millions of scientific papers published every year, sets of challenges and potential directions for addressing them grow rapidly. A striking recent example is that of literature pertaining to the COVID-19 pandemic \cite{wang2020cord}, which exploded in unprecedented volume with researchers from across diverse fields exploring the many facets of the disease and its societal ramifications. 
As the pandemic continues worldwide, it is especially urgent to provide scientists with tools for staying aware of advances, problems, and limitations faced by fellow researchers and medical professionals, and of emerging hypotheses or early indications of potential solutions. 

Unfortunately, due to the immense scale and siloed nature of the scientific community, it can be difficult for researchers to keep track of their own specialty areas, let alone discover relevant knowledge in areas outside their immediate focus \cite{hope2017accelerating,hope2020scisight,hope2021mechanisms, portenoy2021bridger}. This can result in poor awareness of failures or limitations reported in recent studies, wasting redundant resources and leading to clinical decision-making uninformed about shortcomings of interventions \cite{CHALMERS2014}. Disturbingly, there have been many cases where problems in treatments had been reported but not picked up by sectors of the clinical community \cite{IgnoringLimitationsClarke2013ManyRO, IgnoringLimitationsRobinson2011ASE, IgnoringLimitationsCooper2005TheUO} leading to higher rates of morbidity and mortality \cite{KereTXAMetaAnalysis, CostsInLives2005InfantSP, CostsInLivesSinclair1995MetaanalysisOR}. 

Our goal is to bolster the ability of researchers and clinicians to \textbf{keep track of difficulties, limitations and emerging hypotheses}. This could help clinical decision making be well-informed, accelerate innovation by surfacing new opportunities to work on, inspire new research directions, and match challenges with potential solutions from other communities \cite{hope2020scisight}. In the face of challenging medical scenarios, such as the rise of a novel virus or situations where standard treatments fail, rapidly finding reports of similar challenges and directions to address them could have dramatic effect \cite{longhurst2014green}. Finally, at the macro level, this ability could assist policymakers and funding agencies (e.g., NIH, NSF) seeking to identify important challenges and promising directions to prioritize research programs; in times of crisis this process needs to be done rapidly but demands substantial human effort.

\begin{figure*}
    \centering
    \includegraphics[width=\linewidth]{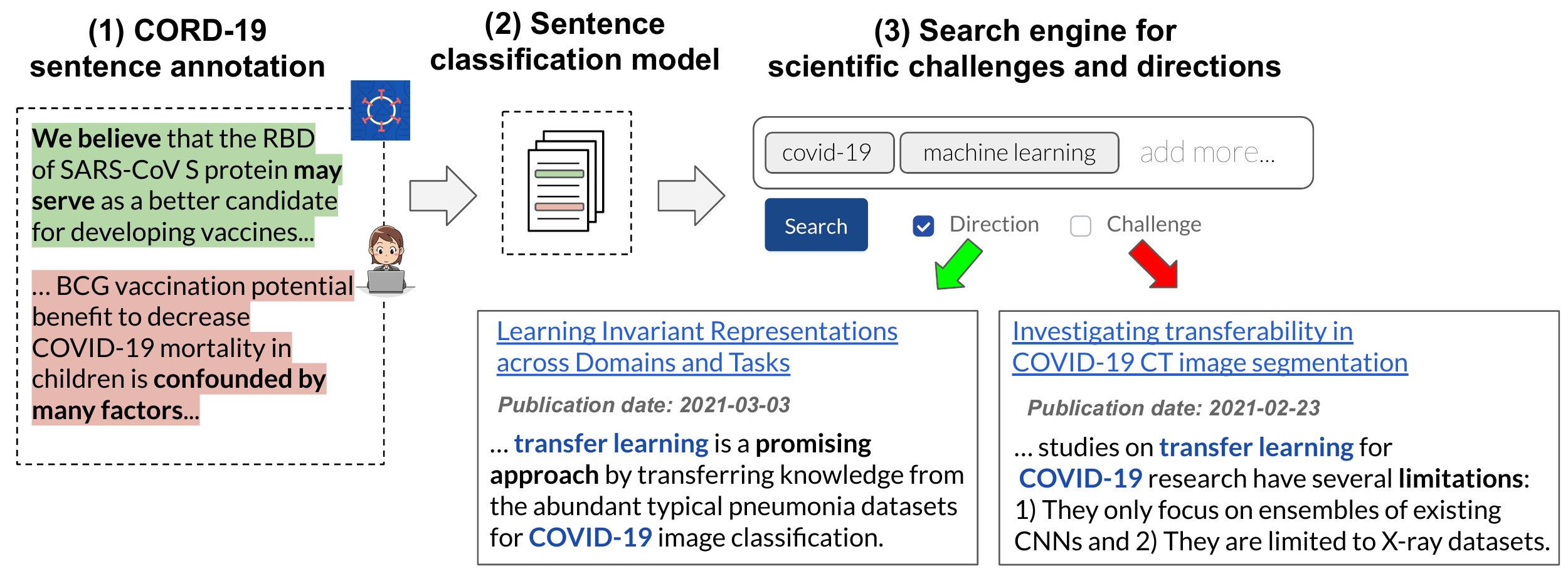}
    \caption{Overview of our system. (1) We collect expert annotations of sentences mentioning challenges and directions from across the CORD-19 corpus. (2) We train a sentence identification model on this data and apply it to the full corpus to extract high-confidence sentences. (3) We build a search engine indexing challenges and directions in COVID-19 literature, allowing users to search for entities and retrieve sentences with their contexts.}
    \label{fig:overview}
\end{figure*}

To address this problem and facilitate discovery of scientific knowledge, we make the following key contributions: 

\begin{itemize}
    \item \textbf{Novel Task: Extraction and Search of Scientific Challenges and Directions.} We define semantic categories for `challenges’ and `directions’ that generalize across many types of difficulties, limitations, flaws and hypotheses or potential indications that an issue is worthy of investigation. We focus on COVID-19 literature as the main test bed for our task, as it is known to be highly interdisciplinary \cite{hope2021mechanisms} with research in many different fields (e.g., AI, climatology, engineering, economics) and relates to a global emergency that urgently demands tools to help researchers and clinicians keep track of challenges and new opportunities.

    \item  \textbf{Expert-Annotated Dataset, Publicly Released.} We collect and publicly release a resource of 2.9K expert-annotated texts from full-length COVID-19 papers, labeled by experts for challenges and directions with high inter-annotator agreement. We use the data to train multi-label sentence classification models that achieve high accuracy scores. We analyze model errors, discovering that contextual information can both help and harm results. Based on this finding, we explore a simple technique that integrates multiple ways of encoding context.
    
    \item \textbf{Novel Scientific Search Engine For Researchers and Clinicians.} We build a novel public search engine that indexes challenges and directions. We apply a model trained on our dataset and apply it to the full corpus of 550K COVID-19 papers to build an index of scientific challenges and potential directions. We create a search engine that allows users to search for combinations of entities (e.g., names of drugs, diseases, etc.) and retrieve challenge/direction sentences that mention them. 
        
    \item \textbf{Evaluating Generality: Zero-Shot Generalization across Biomedicine and AI.} We demonstrate zero-shot generalization, obtaining a high MAP of over 95\% when applying the model trained on COVID-19 papers to a broader corpus in the general biomedical domain, and to AI papers in computer science. This indicates the potential value of our resource beyond COVID-19, such as for future pandemics or crises, or for helping AI researchers handle the explosion of research in this area.
    
    \item \textbf{Evaluating Utility: User Studies with Researchers.} We conduct studies measuring utility. First, we evaluate the system's ability to help researchers with diverse backgrounds discover challenges and directions for a given query (e.g., directions in \textsl{drug discovery}). This could also be important for researchers looking into a new area, e.g., AI researchers seeking biomedical problems (Fig. \ref{fig:overview}). Second, we recruit nine \emph{medical researchers working on COVID-19} in clinical practice and research. These users often require finding information on challenges and directions, during research or treatment planning. In both experiments, totalling 19 researchers and over 70 distinct queries, our prototype outperforms PubMed, the most widely used biomedical search tool, in both quality and utility for discovery of challenges and directions.

\end{itemize}

\section{Task Overview \& Definitions} \label{sec:task}

\subsection{Challenge and Direction Identification}

The CORD-19 corpus \cite{wang2020cord} curates literature on COVID-19 and related diseases. With many thousands of papers, keeping track is generally hard, and mapping the landscape of scientific challenges and directions to address them is even harder. While ``grand'' challenges such as designing therapies and handling novel virus variants are broadly known, research focuses on \textit{fine-grained} specific challenges, e.g., difficulties in functional analysis of specific viral proteins, or shortcomings of a specific treatment regime for children. Each challenge, in turn, is associated with potential directions and hypotheses.\footnote{While many papers discuss future directions in their concluding section, our task involves capturing all mentions of directions/hypotheses/speculations/early indications appearing throughout \emph{full} paper texts (e.g., in experimental analysis sections).}

We present a novel task of automatically identifying sentences in papers that clearly state \emph{scientific challenges and directions}. We consider the multi-label classification setting, where for a given sentence $\mathcal{X} = \{w_1, w_2, \ldots, w_T\}$ with $T$ tokens, our goal is to output two labels $\mathcal{Y} = \{c, d\}$, where $c$ and $d$ are binary targets indicating if the sentence mentions a challenge/direction, respectively. Additionally, we are also given \emph{context} sentences surrounding $\mathcal{X}$: $(\mathcal{X}_\text{previous},\mathcal{X}_\text{next})$, for the previous and next sentences, respectively, which could be used as further input to models. The multi-label setting allows us to capture that in many cases, sentences refer to both challenges and directions at the same time (see Table \ref{tab:labels_examples}). At a high level, our labels are defined as follows.

\begin{itemize}
    \item \textbf{Challenge:} A sentence mentioning a problem, difficulty, flaw, limitation, failure, lack of clarity, or knowledge gap.
    \item \textbf{Research direction:} A sentence mentioning suggestions or needs for further research, hypotheses, speculations, indications or hints that an issue is worthy of exploration.
\end{itemize}

These categories allow us to capture important information for scientists that is not captured by existing resources (see §\ref{sec:related}). As part of data annotation we provide annotators with richer explanations and examples of each label (see §\ref{subsec:datacol}) to make these definitions more concrete. Figure \ref{fig:overview} shows examples for each category (also see Table \ref{tab:labels_examples} in Technical Appendix \ref{sub:examples} for more discussion.

Many cases of challenges and directions are non-trivial for both humans and machines to identify. We demonstrate two main types of difficulties (see more discussion in Technical Appendix \ref{subsec:error_analysis}) --- cases of potentially misleading keywords, and cases where deep domain knowledge or context may be required.
\noindent \begin{itemize}
    \item \textbf{Misleading keywords}. Consider the following sentence: \emph{``The 15-30 mg/L albumin concentration is a critical value that could indicate kidney problems when it is repeatedly exceeded''}. This text mentions a diagnostic measure that is an indicator of a problem, rather than an actual problem. This is one example out of many other potentially misleading cases, such as cases where a negative outcome occurs to an entity we wish to harm (e.g., ``the viral structural integrity is destroyed'').
    \item \textbf{Context and domain knowledge}. \emph{``BV-2 cells expressed Mac1 (CD11b) and Mac2 but were negative for the oligodendrocyte marker GalC and the astrocyte marker GFAP.''} Deciding whether this sentence contains a challenge is highly non-trivial, since it requires more context and deep domain knowledge to understand whether this outcome is problematic or not.  
\end{itemize}

\section{Data Collection and Models}\label{sec:experiments}

We now describe our approach for identifying mentions of challenges and directions, starting with collecting expert annotations needed to train and evaluate models on our task.

\subsection{Data}

\subsubsection{Data Collection \& Annotation}
\label{subsec:datacol}

We recruited four expert annotators with biomedical and bioNLP backgrounds to annotate sentences sampled across CORD-19. Annotators were given detailed annotation guidelines \footnote{Annotation guidelines are available in our code repository.} and had a one-hour training session for reviewing the guidelines and discussing more examples. The guidelines included
simple explanations of challenges and directions along with introductory examples.  
We sampled sentences from full-text papers, aiming to capture diverse, fine-grained challenges/directions that often do not appear in abstracts. The subset of full-text papers in CORD-19 numbers roughly 180K papers with around 25 millions sentences.\footnote{We use a snapshot of CORD-19 from 08-02-2021.} We also provide surrounding sentences around the target sentence as  context.

Randomly sampling sentences for annotation is highly unlikely to lead to enough challenge/direction cases. To increase this likelihood, two annotators curate 280 keywords or phrases with affinity to one of the two categories.\footnote{Our list of keywords is available in our code repository.} Sentences mentioning at least one keyword (lemmatized) are upsampled.  For example, words such as \emph{unknown, limit, however} provide weak signal indicating a potential mention of a challenge; words like \emph{suggest, future work, explore} are weak indicators of a direction. To expand the list further, annotators made use of SPIKE \cite{SPIKE-taubtabib2020} which also has a vocabulary explorer that allows browsing keywords similar to an input term. Overall, the 280 keywords covered around a third of sentences in CORD-19, demonstrating their breadth. We note that for most keywords context can completely change their meaning; for instance, ``limit'' can appear in the context of ``we limit the discussion'' which has no relation to challenges. Our set of terms with weak correlation to the label (e.g., the word \emph{may} that very weakly relates to directions) favors high recall rather than precision.

Finally, to further increase coverage, we sampled at random roughly a quarter of sentences from the remaining sentences that did \emph{not} contain any of the keywords, obtaining in total 3000 sentences. We filter sentences that are not in English, mostly numeric/mathematical, or that are very short/long (often due to PDF parsing issues), resulting in 2894 sentences and their surrounding contexts, from 1786 papers.

\paragraph{Annotator agreement:} 60\% of the sentences were labeled by all annotators\footnote{Final labels selected by majority vote, with ties (fewer than 100 cases) adjudicated by a member of the research team.}, with high average pairwise agreement. Following common practice we measure micro-F1 and macro-F1, treating labels from one annotator as ground-truth and the other as predicted, obtaining 85\% for challenges and 88\% for directions for micro-F1, and 84\% and 82\% for macro-F1. Positive label proportions are 39.66\% and 22.74\% for challenges/directions, respectively. We create a train/dev/test stratified split of 40\%/10\%/50\%(Table §\ref{tab:data_breakdown}), splitting by distinct \emph{papers}. We opt for a large, diverse test set for model evaluation \cite{Card2020WithLP}. The sampled sentences originate from papers published in 1108 journals.
\subsubsection{A note on crowdsourcing.} 
We also attempted crowdsourcing to scale the collection process.\footnote{Using the Appen platform~\url{https://appen.com/}.} However, despite multiple trials and strict quality assurance, the nuanced nature of the task was found to be difficult for crowd workers, especially due to false negatives.

\begin{footnotesize}
\begin{table}[t]
\centering
  \begin{tabular}{@{}lrrrr@{}}
    \toprule
    Labels&Train&Dev&Test&All\\
    \toprule
    Not Challenge, Not Direction & $602$ & $146$ & $745$ & $1493$\\
    Not Challenge, Direction & $106$& $25$ & $122$ & $253$\\
    Challenge, Not Direction & $288$& $73$ & $382$ & $743$\\
    Challenge, Direction & $155$& $40$ & $210$ & $405$\\
  \bottomrule
\end{tabular}
\caption{Distribution of labels across data splits. Splits are stratified with no overlap in papers.}
\label{tab:data_breakdown}
\end{table}
\end{footnotesize}

\subsection{Baseline Models}
\label{subsec:model}
    
The classification task at hand is a multi-label sentence classification problem, with the goal of predicting whether a sentence mentions a challenge, a research direction, both, or neither. The definitions of the challenge and direction categories are as described in §\ref{sec:task}. We evaluate a range of baseline models we examine for our novel task.

\begin{itemize}
    \item \textbf{Keyword-based}: A simple heuristic based on the lexicon we curated for data collection (§\ref{subsec:datacol}) --- sentences with a challenge keyword are labeled as challenge, and similarly for direction. 
    \item \textbf{Sentiment}: Challenge statements potentially have a negative tone, and directions are potentially more positive. We score the sentiment of each sentence using an existing tool \cite{loria2018textblob} and classify negative sentiment sentences as challenges and positive ones as directions. 
    \item \textbf{Zero-shot inference}: In zero-shot classification, models predict labels they were not trained on \cite{yin2019benchmarking}. This could be particularly relevant in emerging domains such as COVID-19, where collecting large amounts of labeled data could be prohibitive. We use a language model trained for natural language inference (NLI), letting the model infer whether the input text \emph{entails} the label name. 
    See Appendix \ref{subsub:Zero-shot baseline} for full details.
    \item \textbf{Scientific language models}: We also experiment with fine-tuning language models that were pre-trained on scientific papers. We report results for PubMedBERT-abstract-fulltext \cite{gu2020domain}  which was pre-trained on PubMed paper abstracts \& full texts, and for SciBERT \cite{beltagy2019scibert}, trained on a corpus of biomedical and computer science papers. 
    In addition, we also experiment with a non-scientific language model, RoBERTa-large, which has been shown to obtain excellent results when fine-tuned on scientific texts \cite{Gururangan2020DontSP}.  We also experimented with other language models, with very similar results.
We fine-tune all language models and perform basic hyperparameter search on the development set. See Technical Appendix \ref{subsub:implement} for full reproducibility details.
\end{itemize}

\subsection{Context Modelling Variants}
We also experiment with models motivated by examination of baseline errors (see Technical Appendix \ref{subsec:error_analysis}). Specifically, we find that adding context helps in certain cases: For example, in the sentence \emph{``... the patient had an extreme elevation of procalcitonin without signs of bacterial infection.''} which was misclassified as a non-challenge, adding context helped identify the unexplained elevation as problematic. However, context can also introduce noise (see Table \ref{tab:modelres}). We explore different ways in which the context can affect predictions --- during training, and during inference. In addition to simply fine-tuning PubMedBERT with full context, we explore two main customized approaches.

\paragraph{Hierarchical Attention Network (HAN)  \cite{HAN}}
Recall Section §\ref{sec:task}, where candidate sentences are denoted by $\mathcal{X}$ and their surrounding context by  $\mathcal{X}_\text{previous},\mathcal{X}_\text{next}$. Denote by $\mathcal{X}_\text{context}$ the concatenation: \texttt{[CLS]} $\mathcal{X}_\text{previous}$  \texttt{[SEP]} $\mathcal{X}$  \texttt{[SEP]} $\mathcal{X}_\text{next}$ \texttt{[SEP]}. We compute a weighted average of \texttt{[CLS]} and the first two \texttt{[SEP]} tokens using attention weights, and use this average embedding for final classification. The weights are learned as part of end-to-end training. \footnote{See \citet{HAN} for details about the general framework.} While this model can potentially learn to re-weight the context, it encodes the full $\mathcal{X}_\text{context}$ jointly before this weighting takes place, which can lead to noise propagating early on. We thus also test a different approach we design.

\paragraph{Context Slice + Combine} Let $f_\mathcal{X}(\mathbf{x})$ denote the logits from the classification layer of the PubMedBERT model fine-tuned on $\mathcal{X}$ only, for input text $\mathbf{x}$. Denote by $f_{\mathcal{X}_\text{context}}(\mathbf{x})$ the logits from PubMedBERT fine-tuned using the \emph{full context}. At inference time, we obtain outputs using the following ``slices'' of $f$ and $\mathbf{x}$: (1) $l_1 = f_\mathcal{X}(\mathcal{X})$, (2) $l_2 = f_{\mathcal{X}_\text{context}}(\mathcal{X}_\text{context})$, (3) $l_3 = f_{\mathcal{X}}(\mathcal{X}_\text{context})$, and (4) $l_4 = f_{\mathcal{X}_\text{context}}(\mathcal{X})$. We then average (``combine'') all four, yielding a final pair of logits used for prediction. (1) and (2) are just the models reported in Table \ref{tab:modelres} -- feeding $\mathcal{X}$ as input to PubMedBERT fine-tuned on $\mathcal{X}$, and similarly for  $\mathcal{X}_\text{context}$. (3) and (4) switch between training and inference inputs: in (3)  $f_\mathcal{X}$ takes $\mathcal{X}_\text{context}$ as input during inference, and in (4) $\mathcal{X}$ is fed as input into $f_{\mathcal{X}_\text{context}}$. The reason we include these is to tease apart different ways in which the context may introduce noise or signal, during training using context (3) and during inference (4). We empirically find all four are in agreement in roughly 70\% / 83\% of the cases for challenges / directions; 3 out of 4 agree in 20\% / 11\%, and the rest are tied. This suggests each variant may capture complementary information.

\begin{footnotesize}
\subsection{Results}
\begin{table*}
\centering
  \begin{tabular}{@{}lrrrrrr@{}}
    \toprule
     & \multicolumn{3}{c}{\bf{Challenge}} & \multicolumn{3}{c}{\bf{Direction}} \\
    \toprule
    Model & P & R & F1 & P & R & F1 \\
    \midrule
    Keyword & 0.535 & 0.760 & 0.628 & 0.455 & 0.792 & 0.578  \\
    Sentiment & 0.405 & 0.966 & 0.571 & 0.239 & 0.837 & 0.371 \\
    NLI-Zeroshot & 0.659 & 0.693 & 0.675 & 0.401 & 0.825 & 0.540  \\
    \midrule
    RoBERTa-large & 0.723 (0.042) & 0.824 (0.046) & 0.769 (0.004) & 0.697 (0.065) & 0.825 (0.06) & 0.754 (0.004)  \\
    SciBERT & 0.729 (0.023) & 0.799 (0.03) & 0.761 (0.007) & 0.719 (0.044) & 0.783 (0.043) & 0.749 (0.01) \\
    PubMedBERT & 0.738 (0.018) & 0.804 (0.017) & \textbf{0.770} (0.006) & 0.755 (0.017) & 0.778 (0.015) & \textbf{0.766} (0.006) \\
    \quad +context & 0.716 (0.048) & 0.809 (0.047) & 0.758 (0.007) & 0.701 (0.038) & 0.771 (0.026) & 0.733 (0.01) \\
    PubMedBERT-HAN & 0.671 (0.02) & 0.863 (0.03) & 0.759 (0.01) & 0.674 (0.04) & 0.804 (0.04) & 0.734 (0.001) \\
    \midrule
    Slice-Combine & 0.742 (0.011) & 0.829 (0.012) & \textbf{0.783} (0.004) & 0.732 (0.02) & 0.82 (0.03) &\textbf{0.773} (0.005) \\

    \bottomrule
  \end{tabular}
    \caption{Model Results. The PubMedBERT model fine-tuned on our multi-label classification task performs best. For the neural models we present the average over 5 training seeds where the number in parentheses is the standard deviation.}
  \label{tab:modelres}
\end{table*}
\end{footnotesize}

\smallskip
\noindent \textbf{Classification Results.}
As seen in Table \ref{tab:modelres}, fine-tuned scientific language models outperform the Zero-Shot model, which still does well considering it had no supervision and was pre-trained on non-scientific texts. The sentiment analysis and keyword-based classifiers, both based on large lists of ``positive/negative'' keywords, have good recall but poor precision. The best individual classifier by F1 is PubMedBERT with a binary-F1 of 0.770 and 0.766 on the challenge and direction labels, respectively. The HAN approach was able to increase recall substantially for problems, but at the cost of reduction in precision, leading to overall inferior F1 on par with PubMedBERT+context.  

The Slice-Combine approach leads to an improvement of about one F1 point for both labels over the best individual model (standard error of $1.05 \times 10^{-4}$). In an ablation experiment we compute the averaged logits of $l_1, l_2$ and $l_3, l_4$ \emph{separately}, and also simply ensemble four model runs of fine-tuned PubMedBERT, both leading to inferior results (see Technical Appendix \ref{subsec:expres_app}). Finally, an oracle that selects the best logit $l_1$-$l_4$ for each input based on ground truth labels has F1 of 0.907 and 0.896 for challenges/directions, suggesting much room for future work on adaptive use of context during training and inference. See in-depth analysis of additional model errors in Technical Appendix \ref{subsec:error_analysis}.

\begin{figure}[t]
    \centering
    \includegraphics[width=0.85\linewidth]{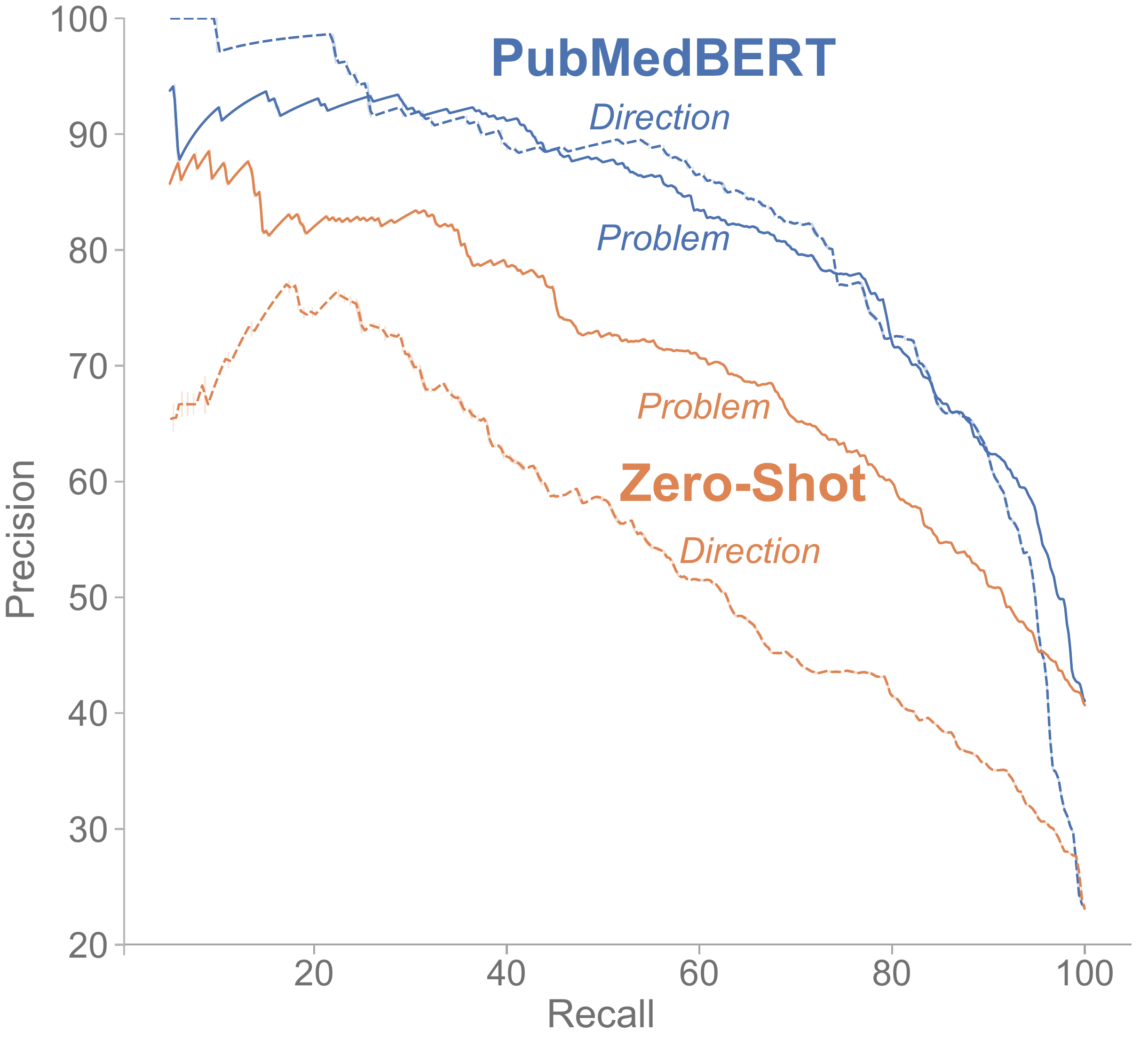}
    \caption{Precision/Recall results for the PubMedBERT model, and the zero-shot model. Precision for PubMedBERT is high for reasonably large values of recall.}
    \label{fig:P@K}
\end{figure}

\paragraph{Precision@Recall} Our primary focus is  a novel search engine application (§\ref{sec:search}). For such applications, it is often more important to have high precision for top retrieved results. We examine precision for a range of values of recall, shown in Figure \ref{fig:P@K}. We observe that for 20\% recall we obtain well over 90\% precision, and for 40\% recall about 90\% precision.

\begin{figure}[t]
    \centering
\includegraphics[width=\linewidth]{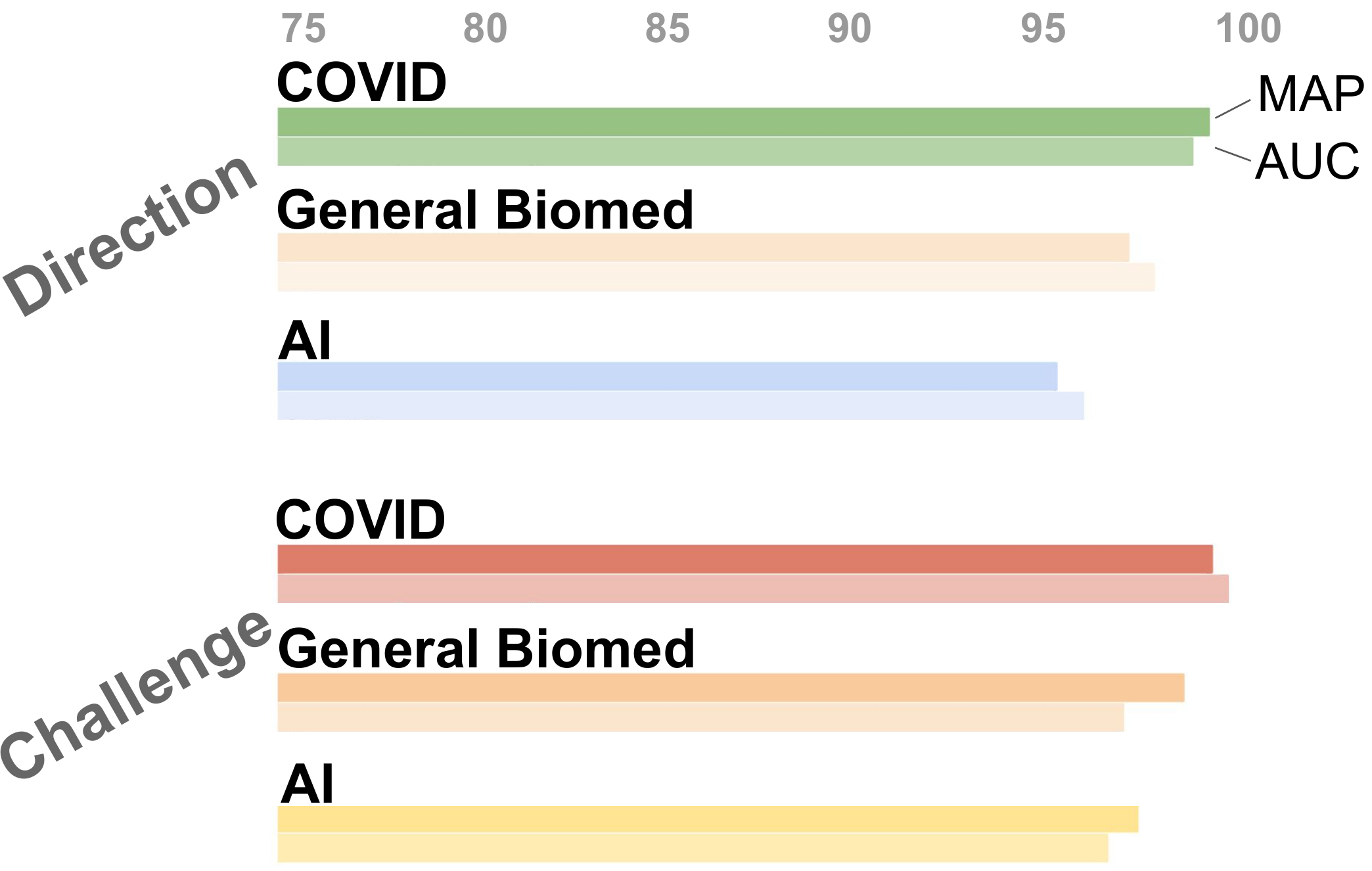}
    \caption{Evaluating predictions beyond our test set. We use a model trained on our data to identify challenges and directions across CORD-19 (denoted by \textit{COVID}), S2ORC (general biomedical papers, denoted \textit{general biomed}) and SciRex (full-text AI papers, denoted \textit{AI}). Accuracy is considerably high. Zero-shot generalization over non-COVID papers, even non-biomedical papers, is encouragingly high, indicating the utility of our resource beyond COVID-19.}
    \label{fig:preds}
\end{figure}

\paragraph{Evaluating predictions across CORD-19} To further ensure quality, we run the PubMedBERT model across all sentences in CORD-19. Out of all sentences indexed in our search engine as either a challenge or a direction, we sample roughly 350 sentences (see Appendix \ref{subsec:expres_app} for details). These sentences are labeled by an expert annotator following the same criteria used to annotate our dataset (§\ref{subsec:datacol}). As shown in Figure \ref{fig:preds}, we obtain very high mean average precision (MAP) of 98\% and area under the precision-recall curve (AUC) of over 97\% for directions, and 97\% / 96\% for challenges. We conclude that for high-confidence challenge and direction sentences indexed in our search engine, accuracy is expected to be overall considerably high. Our test set consists of considerably harder examples, explaining the gap in performance (see discussion in \ref{subsec:error_analysis}).

\smallskip
\noindent {\bf Zero-shot generalization to biomedicine and AI domains}. We explore whether a model trained on our dataset can, with no additional training, generalize to identify challenges and directions in \emph{general} biomedical papers, which we sample from S2ORC \cite{lo-wang-2020-s2orc}, and also AI papers \cite{jain2020scirex}. In total, we sample about 1000 sentences, following the same procedure as described above for CORD-19 sentences (see Appendix \ref{subsec:expres_app} for more details). CORD-19 papers are highly interdisciplinary \cite{hope2021mechanisms,hope2020scisight}, raising the possibility of using our dataset to train models that can be applied to new domains without additional data collection. As seen in Figure \ref{fig:preds}, for directions we obtain MAP and AUC of around 96\% for biomedicine, and around 95\% for AI. For challenges, MAP and AUC reach around 97-98\% for biomedicine and around 96\% for AI. These preliminary results could be explored further in future work.

\noindent {\bf{Training data size.}}
Finally, we also tried training our model on only 10\% of the training set. We obtained average F1 of 0.72/0.71 for challenges/directions. This suggests that a low-resource effort can obtain decent results --- potentially important in emerging scenarios where time is limited.

\section{Search Engine User Studies}\label{sec:search}

We now explore user studies designed to evaluate our framework's utility. First, we explore whether our system can be helpful for quick discovery of challenges/directions. Second, we conduct a study with nine medical researchers working on COVID-19 treatment and research. In total our studies include 19 researchers and over 70 distinct search queries. 

\subsection{Search Engine} \label{subsec:searcheng}

\paragraph{Challenge and direction indexing} We build a search engine that indexes challenges and directions across the entire CORD-19 corpus up to and including August 2021. To build the search engine (see Figure \ref{fig:searchui} in the Appendix for a screen capture), we first apply PubMedBERT to 550K papers, totalling 29M sentences. 180K of the papers are with full text, the rest are abstracts. We then clean poorly tokenized sentences, non-English sentences, very short sentences or texts with latex code.  We classify the remaining sentences leaving 2.2M sentences --- about 950K sentences with high-confidence predictions for at least one of challenge/direction and their surrounding context sentences. We select high-confidence sentences by using a threshold of $0.99$ for both challenges and directions, using a thresholds leading to well over 90\% precision at top-10\% on our test set.

\paragraph{Entity-based indexing} For each sentence in our set of 2.2M, we add another layer of indexing, by extracting entities and linking them to knowledge base entries. This allows us to partially group together all challenges or directions into ``topics'' referring to a specific fine-grained combination of concepts (e.g., \textit{AI + diagnosis + pneumonia}), and facilitate entity-centric faceted search which is known to be useful in scientific exploratory search \cite{hope2020scisight,hope2021mechanisms}.
We extract a range of biomedical entities and link them to a biomedical KB of MeSH (Medical Subject Headings) entities \cite{lipscomb2000medical}. See Appendix \ref{subsub:ent-index} for full details.

In the experiments that follow, we compare our system with a strong real-world system --- PubMed biomedical search engine\footnote{\url{https://pubmed.ncbi.nlm.nih.gov/}}, a leading search site that clinicians and researchers regularly peruse as their go-to tool. While PubMed was not designed to find challenges and directions, no existing tool is; PubMed allows users to search for entities such as MeSH terms, is supported by a KB of biomedical entities used for automatic query expansion, and has many other functions --- and as such is a strong real-world baseline.

\subsection{Challenge/Direction Exploration}
We recruited ten participants with education and experience in medicine, microbiology, public health, molecular, cellular, and developmental biology, biochemistry, chemical \& biological engineering, environmental science, and mathematics. Participants are paid \$50 per hour of work, comparing query results from our system and PubMed. Participants were given guidelines, which include definitions for research challenges and directions with simple examples.\footnote{Full annotation guidelines are included in our code repository.}

Each participant was given twenty queries, split into two sections for challenges and directions, respectively. For each query, participants were asked to find as many research challenges as possible in no more than 3 minutes. The total number of unique queries among the participants is 65. Some examples of queries used for the challenges section include ``antibodies'' and ``inflammation, lung'', with the paired entities being searched jointly; example queries for the directions section include ``telemedicine'' and ``vaccines, technology''. All queries were curated by a domain expert.

\begin{figure}[t]
    \centering
    \includegraphics[width=0.95\linewidth]{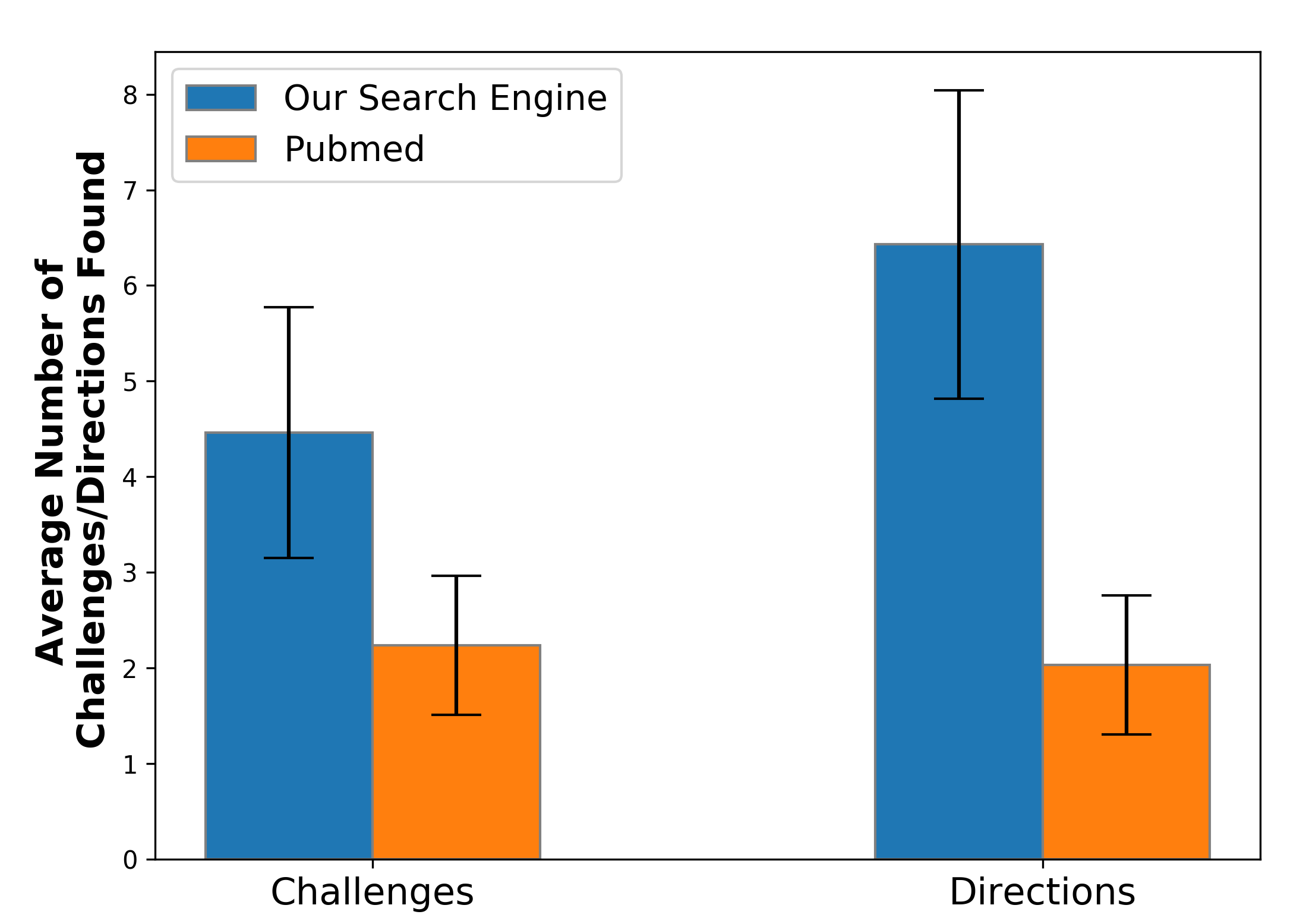}
    \caption{Study participants using our search engine were able to find substantially more challenges and directions they considered useful than with PubMed. Error bars represent 90\% confidence intervals.}
    \label{fig:AverageQuerySentences}
\end{figure}

As seen in Figure \ref{fig:AverageQuerySentences}, our system yielded a greater number of challenges and directions, on average, than the PubMed tool. Users found roughly 4.46 challenges and 6.43 directions per query using our system compared to the 2.24 challenges and 2.03 directions per query found using PubMed  (p-value of .00192 for challenges and .000529 for directions using a paired t-test). For each participant we included 5 challenges and 5 directions that were overlapping across all participants, in order to control and compare between results for the same queries. We find that on average across users, 70.0\% of the query results using our system led to a strictly larger number of challenges discovered than the respective query results using PubMed, and 22\% were ties. For directions, we find a larger gap between the two systems, with 96.0\% of the query results using our system yielding strictly more directions than PubMed, and 2\% yielding ties.

\begin{footnotesize}
\begin{table}[t]
\centering
    \begin{tabular}{lcc} 
 \toprule
 Metric & Chal./Dir. Search & PubMed \\
 \toprule
 Search  & \textbf{90\%} & 48\% \\
 Utility  & \textbf{94\%} & 57\%  \\
 Interface   & \textbf{91\%} & 68\%\\
 Overall  & \textbf{92\%} & 59\%\\
 \bottomrule
 \end{tabular} 
    \caption{Nine medical researchers expressed much higher satisfaction with our system (Chal./Dir.) than PubMed.}
    \label{tab:md}
\end{table}
\end{footnotesize}


\subsection{Evaluation with Medical Researchers}

We now report on an evaluation of our search engine performed with nine medical researchers at a large hospital.\footnote{See Appendix \ref{subsec:med_search_scenario} for a more detailed example scenario where medical researchers need to search for challenges and directions.}
\paragraph{Study.} We recruited nine expert MDs with a wide range of specialization including cardiology, pulmonary and critical care medicine, gastroenterology and general medicine who are actively involved in clinical research both for COVID-19 and specialty areas, and each have over 1000 citations.
Each expert completed randomly ordered search tasks, in the form of challenge/direction queries curated by an expert medical researcher, using both PubMed and our system. For example, experts performed the following search tasks:

\noindent \begin{itemize}
    \item \textbf{Find problems/limitations} related to COVID-19 and each of (1) \emph{hospital infections}, (2) \emph{diagnosis}, (3) \emph{vaccines for children}, (4)  \emph{probiotics and the gastrointestinal tract}.

    \item \textbf{Find directions/hypotheses} related to COVID-19 and each of (1) \emph{mechanical ventilators}, (2) \emph{liver}, (3) \emph{artificial intelligence}, (4) \emph{drug repositioning}.
\end{itemize}

Experts using our UI viewed sentences and their contexts (previous/next sentences). In addition we also displayed metadata such as paper title, date, url. After all search tasks were completed for both systems, experts were given seven-point Likert-scale questions to judge system utility, interface, and search quality. Following \cite{hope2021mechanisms}, we use a standardized Post Study System Usability Questionnaire (PSSUQ) \cite{lewis2002psychometric}, widely used in system quality research, and added questions designed to evaluate search and exploration utility: \emph{overall search accuracy}, \emph{results that are not only relevant but interesting or new}, \emph{finding papers interesting to read}, and \emph{ability to understand and judge each individual result quickly without additional context}. Each question is asked twice, once for PubMed and once for our system, leading to 15$\times$2$\times$6 = 180 responses. 


\vspace{.1cm} \noindent {\bf Results.} Table~\ref{tab:md} shows the average Likert scores (normalized to  [0\%,100\%]). We group questions by three types for brevity. The medical experts strongly prefer our search engine to PubMed (overall average of 92\% vs. 59\%, non-normalized scores of 6.42 vs. 4.14). Across all questions, the majority of the nine MDs assigned our system a higher score than PubMed, at an average rate of 85\% per question. When considering ties, the rate is 92\%. Our system significantly outperformed PubMed across all questions (Wilcoxon signed rank test p-value of $5.409\times10^{-6}$). 

\section{Related Work}\label{sec:related}



\textbf{Scientific information extraction and text classification.} The goal in this line of work is to extract structured information from literature, such as sentence-level classification into categories including objectives/methods/findings \cite{dernoncourt2017pubmed} or extracting entities and relations \cite{li2016biocreative,PICO-wallace, kim2013genia}. Unlike previous work, our labelling schema encapsulates underexplored facets and covers diverse variants of challenges/directions and can help generalize across the interdisciplinary COVID-19 literature \cite{hope2021mechanisms}. Some previous schemas focus on subsets of categories that are subsumed by ours, which is broader in scope and captures aspects missed by other work (e.g., societal hardships)\cite{teufel-2009-towards, liakata-etal-2009-semantic, fisas-2015-discoursive, fisas-2016-multi}. Most previous work was also multi-\textit{class} rather than multi-\textit{label}, which excludes important cases of statements that are both challenges and directions. Importantly, most other datasets are much narrower and limited in size, with the largest relevant corpus consisting of about 10X fewer papers than our own. 

\textbf{COVID-19 IE and search tools.} Recent work includes visualizing COVID-19 concepts and relations \cite{hope2020scisight}, a syntactic search engine \cite{Shlain2020SyntacticSB}, and a search engine for causal and functional relations \cite{hope2021mechanisms}. Ours system is focused on challenges and directions, not captured by existing tools. Recent work \cite{huang2020coda} has used crowd workers to annotate \emph{abstracts} (not full-texts as in this paper) for Background, Purpose, Method, Finding/Contribution. As discussed in §\ref{subsec:datacol}, we find that crowd workers fail on our task, even though recruited with high quality assurance standards.

\section{Conclusion}
We presented methods for extracting scientific challenges and directions from scholarly papers. We collected 3K expert-labeled sentences and their contexts from COVID-19 papers, and used the dataset to fine-tune scientific language models on our multi-label sentence classification task. Our model identifies challenges and directions with high precision, and achieves high zero-shot generalization on general biomedical and AI papers. We used the model to index 950K sentences and build a novel search engine that allows researchers to search for biomedical entities and retrieve sentences mentioning difficulties, limitations, hypotheses and directions. Researchers using our system, including those working on COVID-19, found that our system provided better support than PubMed in terms of utility and relevance. 
In future work, we hope to build more tools to explore and visualize challenges and directions across science. 

\section{Acknowledgments}
Lahav is partially supported by the Edmond J. Safra Center for Bioinformatics at Tel-Aviv University.
Weld's work at the University of Washington is funded by ONR grant N00014-18-1-2193, NSF RAPID grant 2040196, the WRF/Cable Professorship, and AI2.

\begin{quote}
\begin{small}
\bibliography{aaai22}
\end{small}
\end{quote}

\newpage
\clearpage
\appendix

\section{Technical Appendix}
\subsection{More examples}
\label{sub:examples}

Table \ref{tab:labels_examples} shows example sentences for each category. In the first row (\emph{not challenge, not direction}), the example is purely a factual description of a certain tool. In the second row (\emph{not challenge, direction}), the statement mentions a scientific future direction, but there is no associated challenge that is explicitly mentioned. In the third row (\emph{challenge, not direction}), there is a mention of a disease that is difficult to diagnose, but there is no mention of a suggested hypothesis or direction. Finally, in the last row (\emph{challenge, direction}), a medical concern is presented alongside a scientific speculation on the nature of the signaling in the immune system, therefore reflecting both a challenge and a direction. 

\begin{footnotesize}
\begin{table}[t]
\centering
\begin{tabular}{p{22mm}p{50mm}}
\toprule
Labels&Example\\
\toprule
Not Challenge,\newline Not Direction & Nowadays, standard structure-based virtual screening has been routinely implemented in drug discovery to quickly prioritize potential compounds for in vitro activity tests.\\
Not Challenge,\newline Direction & Future studies will focus on comparative sequence analysis between the PST isolates reported herein and global isolates of PST to determine the specific geographic origin(s) for this diverse PST population.\\
Challenge, \newline Not Direction & Outbreaks attributed to acute BVDV infections in feedlot calves have been described previously, although definitive diagnosis is often difficult [18].\\
Challenge, \newline Direction & Thus, both PRRs could be responsible for innate immune signaling during acute DENV infection, perhaps operating in temporally distinct fashion as in WNV infection.\\
\bottomrule
\end{tabular}
\caption{Examples of Challenges and Directions.}
\label{tab:labels_examples}
\end{table}
\end{footnotesize}

\subsection{Baseline models}
\subsubsection{Zero-shot baseline}
\label{subsub:Zero-shot baseline}

We use BART-MNLI-large \cite{lewis2019bart}, a pre-trained NLI model. We find that simply feeding in ``challenge'' and ``direction'' as  label names, or similar variants, performs poorly, likely due to the nuanced complexity of these labels. Instead of using one name, we find that enumerating multiple variants of challenges (e.g., difficulty, limitation, failure) provides better results.
 
We define the following sub-labels enumerating challenges and directions. We take the different variants of challenges that we use in our definition of this label --- [challenge, problem, difficulty, flaw, limitation, failure, lack of clarity, gap of knowledge] --- and similarly for directions ( [direction, suggestion, hypothesis, need for further research,  open question, future work]). Denote the former list by $\mathcal{L}_c$, and the latter $\mathcal{L}_d$.  For each category, we compute the probability of each $l \in \mathcal{L}_c$ ($\mathcal{L}_d$, respectively) and take the maximal value for each set of sub-labels, denoted by $m_c$ and $m_d$. If $m_c \geq 0.9$ we label the sentence as a challenge, and similarly for $m_d$ and directions, using the same threshold. Otherwise, the input is classified as negative.

We briefly examine a few variations on the zero-shot classification baseline, in terms of the class/label names given as input, to study their effect. We use the same binary threshold of 0.9 for all variants.
\begin{itemize}
    \item Class-name: Using only the class names, i.e., ``challenge'' and ``direction'', rather than more fine-grained label names.
    \item Template: Using ``challenge'' and ``future direction'' as part of a template sentence following the approach in \citet{yin2019benchmarking}. Specifically, ``This sentence is about a challenge'', ``This sentence is about a future direction''.
    \item Concatenated: Instead of [challenge, problem, difficulty, flaw, limitation, failure, lack of clarity, gap of knowledge] as standalone inputs, we concatenate them into one string --- ``challenge, problem, difficulty, flaw, limitation, failure, lack of clarity, gap of knowledge''; the same was done for directions.
\end{itemize}

Table \ref{tab:ablations} shows results for these variants. For challenges, the variant reported in Table \ref{tab:modelres} achieves the best results by a large margin. For directions, the variant that concatenates class descriptors into one string does marginally better. 

\subsubsection{Implementation details}
\label{subsub:implement}

For all language models we fine-tune we use the Hugging Face library \cite{wolf2020transformers}\footnote{See also the code in our repository.}. We use hyperparameter tuning with the objective of maximizing the F1-score on the development set using grid search over batch size ([8,16,32]), learning rate ([1e\^-05, 2e\^-05, 3e\^-05, 5e\^-05]) and epochs (maximal value of 25 epochs). We use the Adam optimizer\cite{kingma2017adam} with a dropout rate of 0.3 for all neural models, using a binary cross-entropy (BCE) loss over our two labels. 
For the sentiment analysis model, we tune its threshold on the development set.

\subsection{Experiment Results}
\label{subsec:expres_app}

\subsubsection{Slice-Combine Context Models}

In addition to the experiments reported in the paper, we tested multiple ways to combine information from four variants: (i) apply average or a median on the logits, (ii) majority voting, (iii) log-odds extremization, (iv) training a router model based on the logit differences, (v) running logistic regression with the embedding (final layer) of each of the four input encoders and their logits as features. Aside from the simple averaging, logistic regression was a close runner-up. We explore the average weights the logistic regression assigned to the four context variants. For challenges 0.14, 0.21, 0.21, and 0.35 for (1)-(4) respectively; and for directions 0.32, 0.2, 0.07 and 0.45. Interestingly this suggests that training the model end-to-end with context could be useful even when the context is not available at inference.

\paragraph{Sanity testing the Slice-Combine Context Models} As a sanity check we simply ensemble 4 runs of PubMedBERT, resulting in inferior F1 of 0.772 and 0.764 for challenges/directions, which further indicates the complimentary value of our four context variants beyond simpler ensembling. 

\subsubsection{CORD-19 sentences}
We sample roughly 350 sentences, with higher sampling weight given to high-confidence predictions. About 190 sentences have confidence greater than $0.9$, 130 have confidence lower than $0.5$, with 90 sentences with confidence within the range of $(0.25,0.75)$.
\subsubsection{Biomedical and AI sentences}
We randomly sample 100 papers that did not appear in the CORD-19 corpus to ensure no leakage of information from our training set (we filter with a paper identifier shared by both resources). From these papers, we sample sentences in the same way as above for annotation. The annotator labels 630 sentences: 430 sentences with confidence scores greater than $0.9$ and 200 sentences with scores lower than $0.9$, 150 of those with scores lower than $0.5$. For AI papers, we follow the same procedure, with 300 sampled sentences.

\begin{footnotesize}
\begin{table}[t]
\centering
\small
\setlength\tabcolsep{3.5pt} 
  \begin{tabular}{@{}lrrrrrr@{}}
    \toprule
     & \multicolumn{3}{c}{\bf{Challenge}} & \multicolumn{3}{c}{\bf{Direction}} \\
    \toprule
    Variation & P & R & F1 & P & R & F1 \\
    \midrule
    NLI-Zeroshot & 0.659 & 0.693 & \textbf{0.675} & 0.401 & 0.825 & 0.54  \\
    \midrule
    Class-name & 0.789 & 0.440 & 0.565 & 0.618 & 0.065 & 0.119  \\
    Template & 0.439 & 0.941 & 0.599 & 0.589 & 0.401 & 0.478 \\
    Concatenated & 0.849 & 0.107 & 0.190 & 0.491 & 0.724 & \textbf{0.585} \\
    \bottomrule
  \end{tabular}
    \caption{Zero-Shot Baseline Ablation Results. We provide the baseline different variants of class descriptors for challenges and directions, respectively.}
  \label{tab:ablations}
\end{table}
\end{footnotesize}

\subsection{Error analysis}
\label{subsec:error_analysis}

We study the cases where the best fine tuned model failed to classify sentences correctly. In order to do so we randomly sampled and analyzed roughly 20\% of the false positive and false negative errors across both labels. 

\paragraph{Challenges} The most common error that accounts for a third of wrong predictions (both false positive and negative) is that in some cases deciding whether an outcome is positive or negative requires a more profound understanding of the biomedical entities involved and of the context. For example, the sentence
consider the sentence \emph{``The surprising conclusion of the study was that relative to primary rat Schwann cells undergoing myelination, only 2 cell lines expressed high levels of mRNA coding for myelin proteins and none of the cell lines expressed all of the myelin proteins typically expressed in myelinating Schwann cells.''} The model classifies the sentence as a challenge. However, an expert who read the text concluded that the outcome is non-problematic since with further downstream analysis the mentioned conclusion may represent a more accurate model for future analysis of myelin gene promoters. Conversely, the sentence
\emph{``It is remarkable that the patient had an extreme elevation of procalcitonin without signs of bacterial infection.''}, was not classified as a challenge, but an expert annotator did identify the issue of having a strong biological indicator for a serious condition without a clear explanation for its elevation as problematic. We note that in multiple cases presenting the model with the context aids with these issues. For instance, in the above example, when providing the context which includes a reference to the the risk factor, the prediction flips to the right call. However, as discussed in §\ref{subsec:model}, context can also add noise in some cases. This observation led to our Context Slice+Combine approach described in §\ref{subsec:model}.


The second biggest cause for false positives is sentences that provide a general description of a condition rather than a challenge.
An example is \emph{``The location of the headache might vary depending on which sinuses are affected[...]''}. Such texts are tricky since they are essentially facts about a condition rather than a description of an explicit challenge (e.g., the headache may be trivial to treat).
To make the distinction clearer, consider the text \emph{``Colitis is a chronic digestive disease''}. It presents a definition of a disease, and not an instance of an explicit problem one needs to address. 


The second biggest cause we observe for false negatives is sentences that mention \emph{partial} solutions that can mitigate a problem. For instance, \emph{``[...] the cellular apoptotic process is immediately triggered as an innate defense mechanism in response to infection, but is abruptly suppressed during the middle stage of infection.''}. In this example a defense mechanism is mentioned that can mitigate the problem, but the challenge nonetheless remains. Combined, the above error causes account for roughly 2/3 of the false positives and negatives. Labelers were provided input on how to deal with these issues in annotation, but since they are nuanced, models may require more data or sophistication to classify them correctly.

\paragraph{Directions} The most common cause for error in Direction is the identification of future action items which are general vague suggestions as directions. For example, \emph{``In agreement with the authors of that study, we believe that communication between laboratory specialists and clinicians should be intensified and improved.''}. The example suggests a policy or action that does not constitute a research direction or hypothesis. This error accounts for 60\% of the false positives. 

In terms of recall errors we find that sentences that suggest a hypothesis or other more implicit research directions account for roughly 50\% of the errors. For example \emph{``Persistent viral shedding may indicate different levels of virulence, host immune response and infectiousness''}. The guidelines stipulate that these should be positive since researchers need to verify these directions, and indeed in most cases they are correctly identified, but some instances cause false negatives.

The rest of the errors in challenges or directions were anecdotal rather than systemic.

\paragraph{A note on expected errors in the downstream tasks.} Our data contains an over-sampled proportion of tricky keywords (e.g., the word ``hard'' appears in both ``hard material'' and ``hard task''), and thus we expect fewer errors in the search application task (see also Figure \ref{fig:preds}). In addition, in our downstream task of search we rank results by prediction confidence, and precision for high values of confidence is high even on our harder test set (Figure \ref{fig:P@K}). Indeed, Figure \ref{fig:preds} shows that predictions appear to have high overall accuracy in a sampled set. 

\subsection{Entity-based Indexing}
\label{subsub:ent-index}

We employ the SciSpacy library \cite{neumann2019scispacy} to extract entities using five different NER models: one trained on MedMentions \cite{mohan2018medmentions} (a dataset with general mentions of UMLS \cite{bodenreider2004unified} entities covering a wide range of concepts), and four trained on more specialized sources (CRAFT \cite{bada2012concept}, JNLPBA \cite{kim2004introduction}, BC5CDR \cite{li2016biocreative}, BIONLP13CG \cite{kim2013genia}). Each entity is then automatically linked to a biomedical KB of MeSH (Medical Subject Headings) entities \cite{lipscomb2000medical} using SciSpacy's entity linking functionality that performs character-trigram matching on MeSH entity names and aliases. We filter for high-confidence linked entities,\footnote{Using a threshold of 0.9.} and for entities that appeared in at least 10 sentences, then selecting the top 30K unique entities to be indexed by our search engine. At search time, we match user queries to MeSH aliases with an autocomplete dropdown for users to select from as they type. After one entity is selected, the user can search for more from a narrower list of entities that co-occur with it.

\subsection{Search UI}
\begin{figure*}
\centering
    \includegraphics[width=\linewidth]{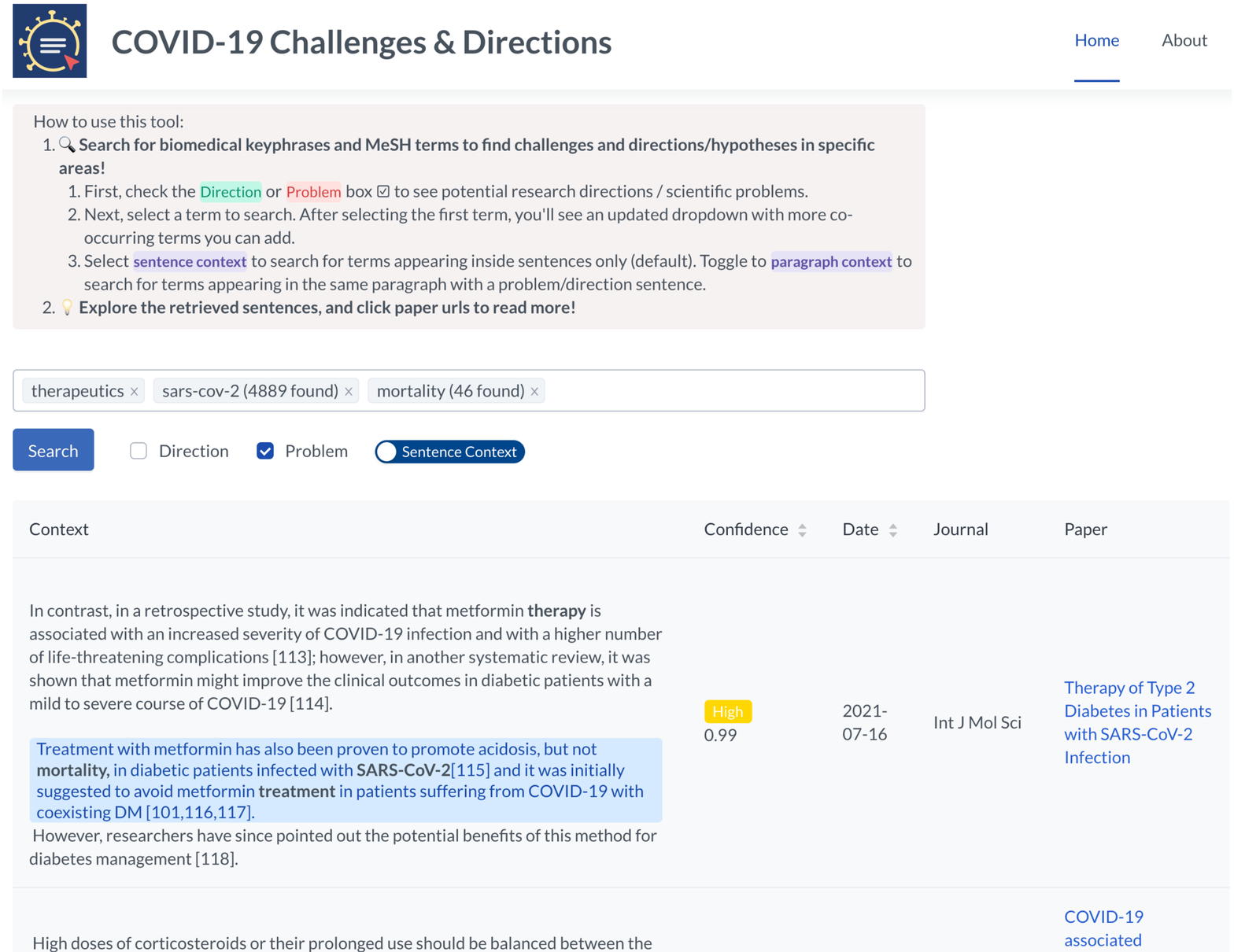}
    \caption{Screen capture of our search user interface.}
    \label{fig:searchui}
\end{figure*}

Figure \ref{fig:searchui} shows a screen capture of our search user interface.



\subsection{Medical Search Scenario}
\label{subsec:med_search_scenario}

In addition to the motivation discussed in the Introduction, we briefly discuss in more detail how a search engine for challenges and directions could help medical researchers when conducting literature reviews. Many research ideas come to MDs with a challenge they perceive during clinical care. If they are unable to find a solution to the problem based on prior experience, they then search for the available scientific literature for possible guidance. When no sufficient guidance is found, research projects are often commenced, starting with literature search which often involves understanding and mapping out associated challenges and directions to help with formulating research questions. Physicians and trainees still spend a significant amount of time doing this form of literature search for fine-tuning their research question. If such a process can be simplified with automation, it could potentially cut down the time and effort needed to formulate and narrow down research questions.


\end{document}